\documentclass[letterpaper, 10 pt, conference]{ieeeconf}  

\IEEEoverridecommandlockouts                              

\usepackage{graphicx} 
\usepackage[table,xcdraw]{xcolor}
\usepackage{multirow}
\usepackage{makecell}

\usepackage[utf8]{inputenc}
\usepackage{todonotes} 
\usepackage{amsmath,amsfonts,booktabs,cite} 
\usepackage{siunitx,textcomp} 
\usepackage[ruled,vlined]{algorithm2e}
\usepackage[hidelinks]{hyperref} 
\usepackage{multicol}
\usepackage{epstopdf} 
\usepackage{subcaption}
\usepackage{float}
\let\labelindent\relax
\usepackage[inline]{enumitem} 
\usepackage[nolist]{acronym} 
\usepackage{colortbl} 	    
\newacro{bn}[BN]{Batch Normalization}
\newacro{relu}[ReLU]{Rectified Linear Unit}
\newacro{adam}[Adam]{Adaptive Moment Estimation}
\newacro{ai}[AI]{Artificial Intelligence}
\newacro{dl}[DL]{Deep Learning}
\newacro{dnn}[DNN]{Deep Neural Network}
\newacro{cnn}[CNN]{Convolutional Neural Network}
\newacro{bnn}[BNN]{Bayesian Neural Network}
\newacro{mc}[MC]{Monte-Carlo}
\newacro{dof}[DOF]{Degrees-Of-Freedom}
\newacro{ods}[USN]{Uncertain Shape Network}
\newacro{od}[SN]{Shape Network}
\newacro{gp}[GP]{Gaussian Processes}
\newacro{mcmc}[MCMC]{Markov Chain Mote Carlo}
\newacro{psdf}[p-SDF]{probabilistic Signed Distance Function}
\newacro{gpis}[GPISs]{Gaussian Process Implicit Surfaces}
\newacro{va}[V]{Varley}
\newacro{spa}[SPA]{Soft Pneumatic Actuator}
\newacro{fsr}[FSR]{Force Sensitive Resistive}
\newacro{pwm}[PWM]{Pulse Width Modulation}
\newacro{rms}[RMS]{Root Mean Square}

\newacro{pi}[PI]{Proportional-Integral}
\newacro{pid}[PID]{Proportional-Integral-Derivative}
\newacro{bic}[BIC]{Bayesian Information Criterion}
\newacro{fem}[FEM]{Finite Element Method}
\newacro{ols}[OLS]{Ordinary Least Squares}

\newcommand{\figref}[1]{\hyperref[#1]{Fig.~\ref*{#1}}}
\newcommand{\tabref}[1]{\hyperref[#1]{Table~\ref*{#1}}}
\newcommand{\secref}[1]{\hyperref[#1]{Section~\ref*{#1}}}
\newcommand{\algoref}[1]{\hyperref[#1]{Algorithm~\ref*{#1}}}

\newcommand{\ra}[1]{\renewcommand{\arraystretch}{#1}}

\newlength\myindent
\def\ie{\textit{i.e.,}}
\def\eg{\textit{e.g.,}}
\def\etal{\textit{et al.}}

\def\rbo{RBO Hand 2}
\def\dlr{DRL Soft Hand}
\def\ourhand{our soft hand}

\definecolor{significant}{RGB}{217,217,217}
\definecolor{verysignificant}{RGB}{255,255,255}

\title{\LARGE \bf
Safe Grasping with a Force Controlled Soft Robotic Hand
}

\author{Tran~Nguyen~Le, Jens~Lundell and Ville~Kyrki%
\thanks{This work was supported by the Strategic Research Council at Academy of 
Finland, decision 314180.}
\thanks{T.~Nguyen~Le, J.~Lundell and V.~Kyrki are with School of Electrical 
Engineering, Aalto University, Finland. 
\texttt{\{firstname.lastname\}{@}aalto.fi}}}

\begin{document}

\maketitle
\thispagestyle{empty}
\pagestyle{empty}

\begin{abstract}

Safe yet stable grasping requires a robotic hand to apply sufficient force on the object to immobilize it while keeping it from getting damaged. Soft robotic hands have been proposed for safe grasping due to their passive compliance, but even such a hand can crush objects if the applied force is too high. Thus for safe grasping, regulating the grasping force is of uttermost importance even with soft hands. In this work, we present a force controlled soft hand and use it to achieve safe grasping. To this end, resistive force and bend sensors are integrated in a soft hand, and a data-driven calibration method is proposed to estimate contact interaction forces. Given the force readings, the pneumatic pressures are regulated using a proportional-integral controller to achieve desired force. The controller is experimentally evaluated and benchmarked by grasping easily deformable objects such as plastic and paper cups without neither dropping nor deforming them. Together, the results demonstrate that our force controlled soft hand can grasp deformable objects in a safe yet stable manner.
\end{abstract}

\section{Introduction}
\label{sec:introduction}

\begin{figure}[t]
  \begin{center}
    \includegraphics[width=0.48\textwidth]{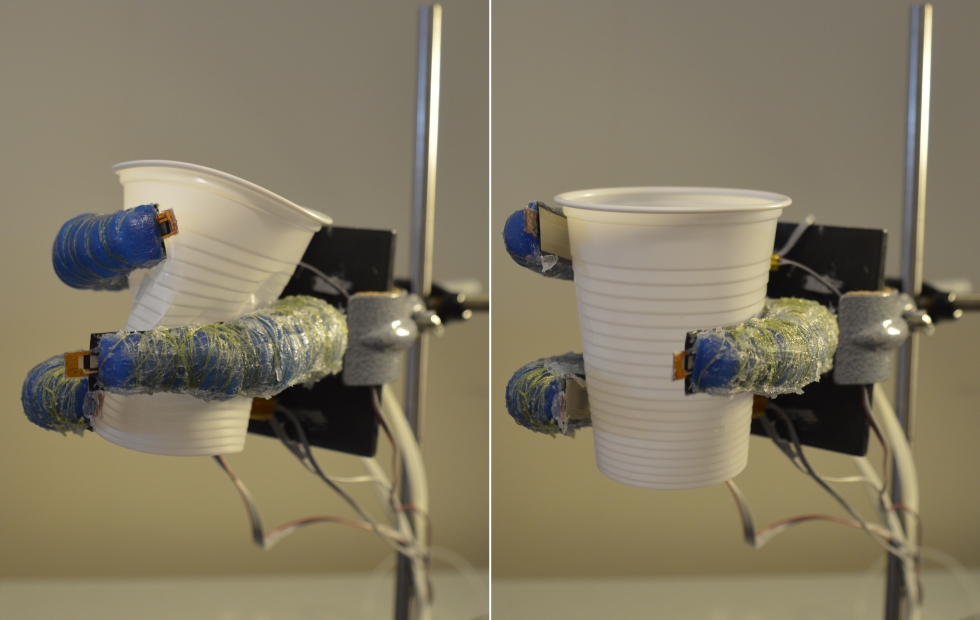}
  \end{center}
    \caption{On the left is shown that grasping a fragile object with a high force will deform it completely while, as shown on the right, using force-feedback to control the grasping force this is no longer the case.}
  \label{fig:resultfigure}
\end{figure}

A stable yet safe grasp requires a robotic hand to apply sufficient force on the target object to not harm it but still immobilize it. Recently, soft hands have been proposed for safe grasping due to their passive compliance. However, as shown in \figref{fig:resultfigure} even such hands can deform objects if the applied force is too high. There is, therefore, a need to control the grasping force also in soft hands. 

Research on soft hands has mainly focused on hand design \cite{schulz_anthrohand_2001,dollar_sdmhand_2010,filip_starfish_2011,deimel_rbo1_2013,deimel_rbo2_2016,galloway_underwater_gripper,zhou_40N_gripper,gecko_design,bryan19,nishimura19}. Only recently have works started to address the integration of sensing capabilities such as position \cite{elgeneidy_bendpredict_2018,chin_deform_19,gerboni_17,soter18} and force sensing \cite{homberg_robustgrsp_2019}. However, the authors found the existing solution for force sensing to be insufficient for contact detection, especially for small objects \cite{homberg_robustgrsp_2019}. In this work, we aim to show that force estimation is possible to allow both contact detection as well as control of contact forces for safe grasping.

To this end, we propose to integrate off-the-shelf resistive bend and force sensors in each finger of a hand to estimate the contact force and feed it into a force controller for each finger. In order to extract reliable force measurements, we propose a data-driven sensor characterization and calibration method for such sensors. Given the calibrated sensors, the real contact force is estimated by subtracting the residual force the sensor senses when bending in free space from the actual force reading. The residual force is also learned. The estimated force is then used in a \ac{pi} controller to control the set-point force for each finger. The complete force controller is experimentally validated and compared to no force controller on a physical grasping experiment where the objective is to grasp deformable and fragile objects in a stable manner without neither destroying nor deforming them.

The main contributions of this work are:
\begin{enumerate*}[label=(\roman*)]
	\item integration of resistive bend and force sensors in a soft hand (\secref{sec:sensing}), 
	\item calibration of the sensors to estimate contact force (\secref{sec:est_contact_force}),
	\item a force controller for the soft hand (\secref{sec:Force_controller}), and
	\item an empirical evaluation of the controller demonstrating that using a force controller deformable objects can be grasped in a stable yet safe manner (\secref{sec:Experiments}).
\end{enumerate*}
\section{Related Work}
Most of the research in soft robotic hands has focused on the hand design itself where the main actuators have been the \ac{spa} \cite{shintake_softgripper_18}. \ac{spa}s are constructed of soft and deformable matter and are pneumatically actuated. Such actuators were used in the pioneering soft hand design by Ilievski \etal{} in 2011 \cite{filip_starfish_2011} where they presented a three-layer starfish inspired soft gripper using embedded pneumatic networks or PneuNets. The multilayer structures, where two active layers were separated by a passive layer, allowed the gripper to perform a wide range of motions, but due to its construction it could only withstand a small amount of pressure resulting in a weak grasping force. 

Recently, alternative soft hand designs that can attain higher grasping forces have been proposed, including the well known anthropomorphic \rbo{} \cite{deimel_rbo2_2016} and \dlr{} \cite{homberg_haptic_2015}. The \rbo{} consists of seven fiber-reinforced pneumatic continuum actuators acting as fingers and palm. Although the anthropomorphic design enabled dexterous grasping, one downside is the negative curvature of the thumb, meaning that the backside of the thumb is the primary contact surface rather than its front side, which makes it problematic to attach sensors to it. Compared to the \rbo{}, the kinematics of the \dlr{}, where two fingers are parallel and opposes the third, allow easy sensor integration and was, therefore, the source of inspiration for the hand proposed in this work. 


Despite the wide range of research in soft hand designs, few works have studied how to incorporate sensing capabilities in such hands. One possible reason for the lack of research in sensorized soft hands is the few existing cheap off-the-shelf sensors that can be integrated into a \ac{spa}. However, recently Elgeneidy \etal{} \cite{elgeneidy_bendpredict_2018} proposed the following sensors for measuring and controlling position of \ac{spa}s: 
\begin{enumerate*}[label=(\roman*)]
	\item conductive elastomer sensors, 
	\item sensors made from liquid metal or
	\item resistive flex sensors.
\end{enumerate*}

Conductive elastomer sensors were used in \cite{petkovic_elastomer_2012} to control the displacement of the soft gripper with an adaptive neuro-fuzzy inference strategy. Unfortunately, the accuracy of such sensors quickly deteriorates as the distribution of carbon particles inside the elastomer material is disturbed by the actuator's repeated deformation. The more durable liquid metal sensors were integrated into a soft gripper in \cite{eGaIn_presence} to detect the presence of grasped objects, and in \cite{morrow_force_2016} to acquire pressure, force, and position data which was subsequently fed to a \ac{pid} controller to control the position of the finger and grasping force. However, to acquire the data, they used the eGain liquid metal sensor \cite{Park_eGaIn_2012} that was fabricated from scratch, which is a time-consuming and error-prone process. Additionally, the force sensor was placed at the tip of the actuator, restricting the force controller to only work with pincer grasps or on large objects. In this work, we target objects of diverse sizes, including small objects and as such the eGain sensor is not applicable.   

The third sensor type mentioned in  \cite{elgeneidy_bendpredict_2018}, the resistive bend flex sensors, are cheap, widely available, and easy to integrate in soft hands. For instance, resistive bend sensors have been used for haptic identification of grasped objects \cite{homberg_haptic_2015} and for predicting and controlling the position, \ie, bending angle of a pneumatic-driven actuator \cite{elgeneidy_bendpredict_2018}. The work in \cite{homberg_haptic_2015} was later improved in \cite{homberg_robustgrsp_2019} by adding a \ac{fsr} sensor to strengthen the grasp by detecting the contact between the hand and objects. They reported, however, that the force sensor provided unreliable data, resulting in extremely poor performance in contact detection, especially on small objects. This problem stems from placing the force sensor only at the tip of the finger, which will not make contact with small objects such as a tennis ball, resulting in no readings. In this work, we also use \ac{fsr}s due to their simplicity and availability but opt for the kinds that measure forces along a wider area as they enable measuring forces even on small objects.
\section{Resistive Based Sensing}
\label{sec:sensing}

\begin{figure}[t]
  \begin{center}
    \includegraphics[width=0.48\textwidth]{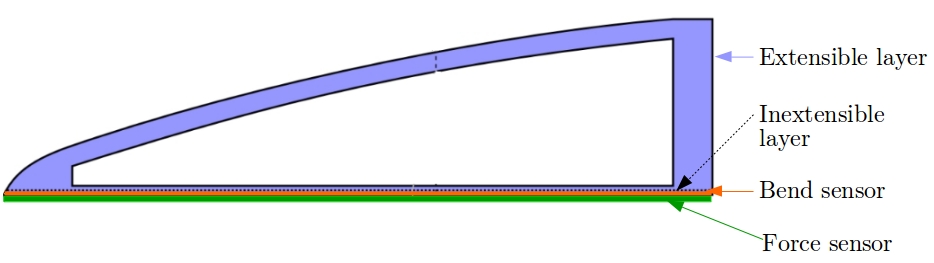}
  \end{center}
    \caption{A cross-sectional view of the actuator embedded with selected sensors.}
  \label{fig:sensorintegration}
\end{figure}

To do complex manipulation tasks, robotic hands need position and force feedback \cite{manipulation_requirement}. To achieve this for soft hands, we propose integrating deformable resistive force and bend sensors onto the finger of the RBO Hand 2 proposed in \cite{deimel_rbo2_2016}. \figref{fig:sensorintegration} shows the sensors integrated into the fabricated actuator. The body of the fabricated actuator is divided into two parts: an extensible and an inextensible layer. 

To keep the bend sensor in place, it was encapsulated in the inextensible layer shown as the orange line in the figure. In contrast to the bend sensor, the force sensor illustrated by the green line in the figure needs to be in contact with the environment to get measurements; thus it was glued directly to the outer surface of the inextensible layer using freshly mixed Dragon Skin 10 silicone. Next, we describe the design and working principle of these sensors.
\subsection{Position sensing} 
\label{sec:pos_sensing}

\begin{figure}[t]
  \begin{center}
    \includegraphics[width=0.48\textwidth]{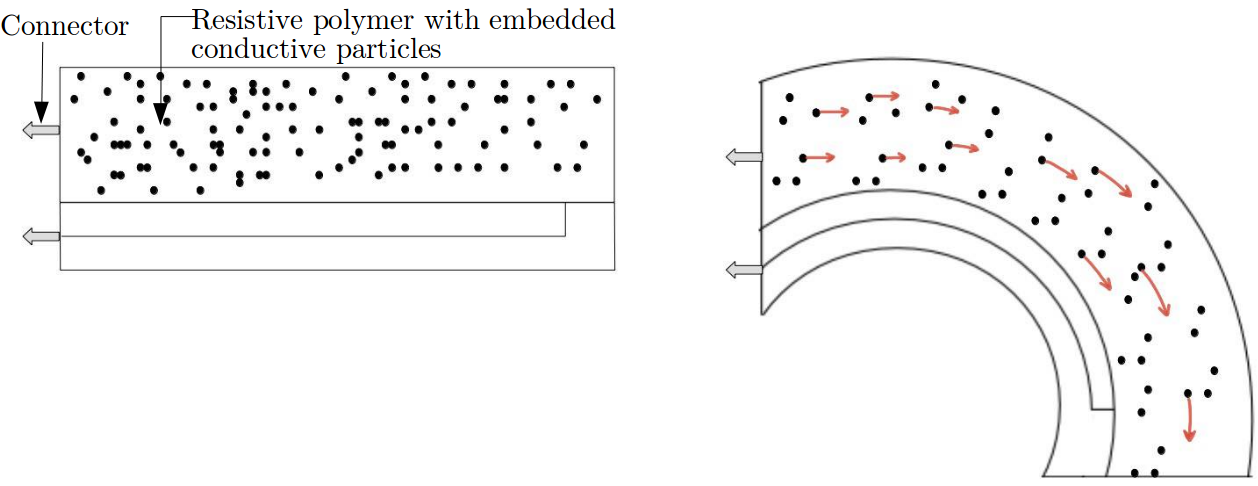}
  \end{center}
    \caption{The left figure shows the design of a resistive flex sensor and its behaviour in straight state. The right figure illustrates the behaviour of the sensor when it is bent in the right direction.}
  \label{fig:flexprin}
\end{figure}

To measure the internal state (position) of \ac{spa}s we used resistive flex sensors (bend sensors)\footnote{Flex Sensor 4.5": \url{https://www.sparkfun.com/products/8606}}. One side of the sensor consists of a layer of polymer ink which is embedded with conductive particles, as shown in \figref{fig:flexprin}. The particles provide the ink with a certain amount of resistance when the sensor is straight. When the sensor bends away from the ink, the conductive particles move further apart as shown in \figref{fig:flexprin} which, in turn, increases the resistance. Once the sensor return to the initial pose, the resistance also returns to the original value. Hence, the change in the resistance can be used to determine the curvature of the sensor.

\subsection{Force sensing}
\label{sec:force_sensing}

To measure the grasping force applied by \ac{spa}s, we use \ac{fsr}s. \ac{fsr}s are designed to measure the presence and relative magnitude of localized physical pressure. The \ac{fsr} used to measure the force in \cite{homberg_robustgrsp_2019} was placed only at the tip of the finger but, given the fact that soft hands typically are developed to mimic human's grasping behaviour, \ie\ wrapping an entire finger around the surface of an object, placing the force sensor at such a location is not optimal for detecting contact between the hand and the object. Based on this, we opted for force sensors that can provide measurements along the body of the whole finger rather than only at the tip. To meet this requirement, we chose a strip \ac{fsr} with rectangle shape\footnote{Force Sensitive Resistor - Long
: \url{https://www.sparkfun.com/products/9674}}. The chosen sensor is a single large sensing taxel with the force sensitivity ranges from $0.01$ to $1000$ N. In addition, the force resolution of the sensor is better than $± 0.5 \%$ of full use force. 

\begin{figure}[t]
  \begin{center}
    \includegraphics[width=0.48\textwidth]{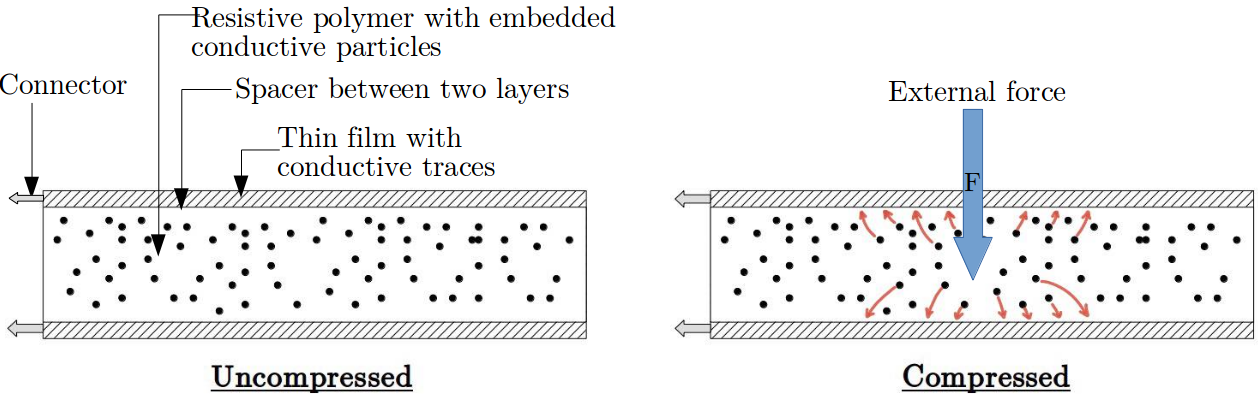}
  \end{center}
    \caption{The left figure shows the design of a strip \ac{fsr} and its behavior when no pressure is applied. When pressure is applied on the sensor, the conductive particles move and make contact with the conductive film, resulting in more conduction paths (lowering resistance). }
\label{fig:fsrprin}
\end{figure}

An \ac{fsr} consists of a resistive polymer layer and a conductive film as illustrated in \figref{fig:fsrprin} and the working principle is similar to that of the bend sensor. Specifically, the spacer placed in between the two layers to avoid contact between them results in a very high resistance when no pressure is applied. When pressure is applied, the resistive polymer starts to make contact with the conductive film, which, in turn, reduces the resistance of the sensor. The stronger the applied force is on the sensor’s active area, the more the resistance between the two terminals drops. The actual applied force is mapped from the measured resistance using the resistance-force relationship graph provided in the datasheet of the sensor\footnote{https://cdn.sparkfun.com/datasheets/Sensors/Pressure/FSR408-Layout2.pdf}.

The design of the \ac{fsr} enables easy contact detection in stationary situations when the sensor is fixed to a surface. However, when the sensor is bent in free space the conductive and resistive polymer layer can come in contact with each other producing \textit{internal force readings} which leads to incorrect contact forces. In the next section, we propose a data-driven calibration method to correctly estimate the contact force by compensating for the internal force.

\section{Data-Driven Contact Force Estimation}
\label{sec:est_contact_force}
\subsection{Internal Force Characterization}
\begin{figure*}
  \begin{center}
    \includegraphics[width=0.95\textwidth]{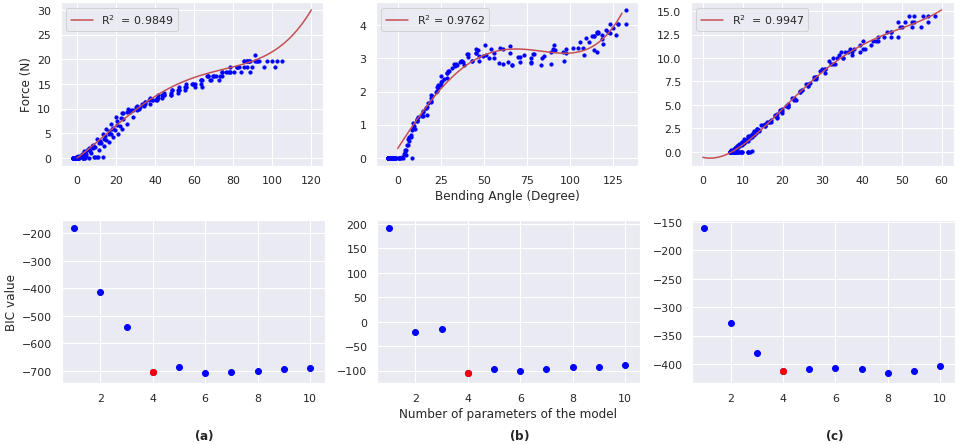}
  \end{center}
    \caption{The upper row (from left to right) shows, as blue dots, the internal force for each finger (from finger 1 to finger 3) at different bending angles along with the fitted polynomial model in red. The bottom row shows the BIC values for each finger.
}
  \label{fig:forceestimator}
\end{figure*}

To characterize the internal force of the force sensor, a finger of \ourhand{} with integrated sensors was actuated in free space with a 60 kPa ramp input. Once the internal pressure in the finger reached 60 kPa, the input was zeroed, and the finger returned to its initial position. The top left plot in \figref{fig:forceestimator} shows the force measurement of the force sensor against the bending angle after 35 repetitions. Theoretically, the force measurements should remain zero as there is no contact between the finger and the environment. However, as seen in the figure, the force reading increases the more the finger is bent. Due to such sensor characteristics, the FSR force measurements of a bent finger are incorrect. Thus, removing the effect of the internal force from the sensor readings is crucial to enable force control of the fingers.  

\subsection{Contact Force Estimation}
\label{sec:calib_contact_force_est}
We model the actual contact force $F_{c}$ for the finger by the difference between the measured force $F_{m}$ and the internal force $F_{i}$
\begin{equation}
    F_{c} = F_{m} - F_{i}.
\label{contactforce_equation}
\end{equation}
Thus, to estimate the actual contact force $F_c$, we 1) learn the internal force $F_{i}$ caused by bending in free space and 2) subtract the estimated internal force from the measured force.

We cast the learning of the internal force model $F_{i}$ for each finger as a polynomial regression problem since the relationship between internal force and bending angle of the finger is non-linear (\figref{fig:forceestimator}). Thus the internal force is expressed as
\begin{equation}
     F_{i}(x) = \sum_{d=0}^{D} w_{d} x^{d}=\mathbf{w}^{\text{T}}\mathbf{x},
\label{eq:polymodel}
\end{equation}
where $\mathbf{w}=[w_1,\dots,w_D]$ are the parameters to learn, $x$ the bending angle of the finger, and $D$ the maximum degree of the polynomial function. We estimate the model parameters by first mapping the bending angle $x$ to a higher dimensional feature space using the feature map $\phi(x)= [x^{d},\dots,1]^{\text{T}}$, resulting in a transformed feature vector $\mathbf{x}=[x^{d},\dots,1]^{\text{T}}$, and then applying ordinary least squares so that the optimal parameters $\mathbf{w}^*$ can be found by $\mathbf{w}^{*} = (\mathbf{X}^T \mathbf{X})^{-1}\mathbf{X}^T \mathbf{F}_i$. 

To choose the model complexity, that is the maximum degree $D$ of the polynomial function, we used the \ac{bic} \cite{bic}
\begin{equation}
    \text{BIC} = \ln(n) k - 2 \ln(\hat{L}),
\end{equation}
where $\hat{L}$ is the maximized value of the likelihood of the model, $n$ is the number of observations, and $k$ is the number of parameters. The model with smallest BIC value is considered the best. 

\section{FORCE CONTROLLER}
\label{sec:Force_controller}
To control the grasping force of the soft hand, we propose using a discrete \ac{pi} controller 
\begin{equation}
    u_n = K_p e_n + K_i T \sum_{k=1}^{n} e_k,
    \label{eq:force_controller_discrete}
\end{equation} 
where $K_p$ and $K_i$ are the proportional and integral terms respectively, $T$ is the sampling period, and $e_n$ is the force error between the target value and measured value at the n-th moment of sampling. The output $u_n$ is a \ac{pwm} signal. The reason why we chose to use a \ac{pi} controller instead of a \ac{pid} controller is that the derivative action is sensitive to noise. As the force measurement of the force sensor is usually noisy, the derivative action is neglected.



\section{EXPERIMENTS AND RESULTS}
\label{sec:Experiments}

The main questions addressed experimentally were:
\begin{enumerate}
    \item What is the accuracy of the contact force estimation?
    \item What is the accuracy of the force controller?
    \item Does the force controlled soft hand enable safe grasping?
\end{enumerate}
In order to provide justified answers to these questions, we conducted three experiments. The first experiment examines the contact force estimation accuracy, the second one the proposed force controller accuracy while the third evaluates safe grasping in terms of grasping deformable objects without neither damaging nor dropping them.

\subsection{Experimental setup}
All the experiments were evaluated on a soft robotic hand embedded with the bend and force sensors mentioned in \secref{sec:sensing}. For the hand design, we used the same finger structure as \rbo{} \cite{deimel_rbo2_2016} but opted for a kinematic structure similar to the \dlr{} \cite{homberg_haptic_2015} as it allows power grasping. As a result, \ourhand{} consisted of three fingers, in which two fingers are on one side (we call them \textit{finger 1 and finger 2}), and one finger is on the opposite side (\textit{finger 3}). The final soft hand is shown in \figref{fig:resultfigure}. 

To pneumatically control the hand, we used the soft robotics toolkit controller platform\footnote{Fluidic Control Board, \url{https://softroboticstoolkit.com/book/control-board}}. The input pressure was regulated with \ac{pwm}, and the control rate was 60 Hz. However, at such a high control rate, the high-speed switching of \ac{pwm} caused the hand to vibrate, which consequently induced noise into the force readings. To decrease the noise, we implemented the pneumatic low pass filter proposed in \cite{memarian_lowpass_2015}.

\subsection{Estimating contact force}
Our hypothesis is that the internal force is correlated to the curvature of the finger. To test this, we gathered force data at different bending angles for each finger. The data in \figref{fig:forceestimator} shows the force measurements for each finger at different bending angles. Although the data for each finger, such as bending angle range differ from each other due to the manual fabrication process, we can clearly see the correlation between the bending angle and the force. Next, we fitted a separate polynomial model \eqref{eq:polymodel} to the data for each finger. According to the \ac{bic} values presented in \figref{fig:forceestimator}, a fourth-degree polynomial fitted the data for each finger best, which is also indicated by the high $R^2$ values. 

\begin{figure}[t]
  \begin{center}
    \includegraphics[width=0.45\textwidth]{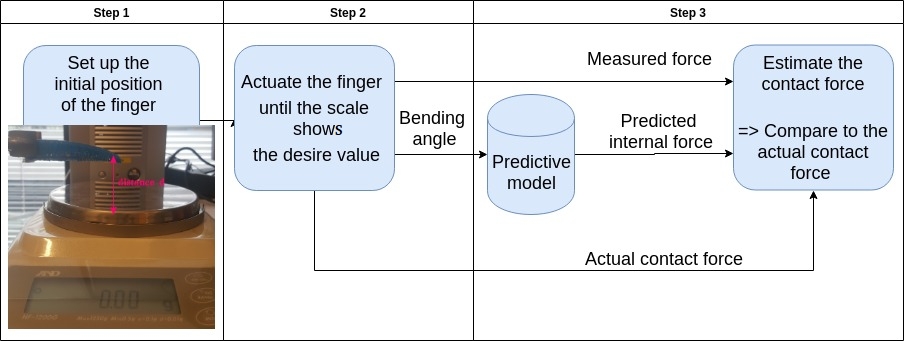}
  \end{center}
    \caption{Steps of the experiment conducted to evaluate the accuracy of the estimated contact force using the proposed approach.}
  \label{fig:estimation_acc}
\end{figure}

Next, using the models to predict the internal force, we evaluated the accuracy of the estimated contact force in \eqref{contactforce_equation}. The steps of the experiment are illustrated in \figref{fig:estimation_acc}. The accuracy was measured by first actuating one finger to press against a scale until the reading on the scale reached a predefined target set to an arbitrary value such as 200 gram (approximately 2 N) in this case. Once the target was reached, the actual contact force was estimated by subtracting the predicted internal force from the force measurement obtained from the sensor. In addition, to evaluate the accuracy of the proposed method at different finger configurations, the distance $\mathbf{d}$ between the finger and the scale was increased after each experimental cycle.

\begin{figure}[t]
  \begin{center}
    \includegraphics[width=0.45\textwidth]{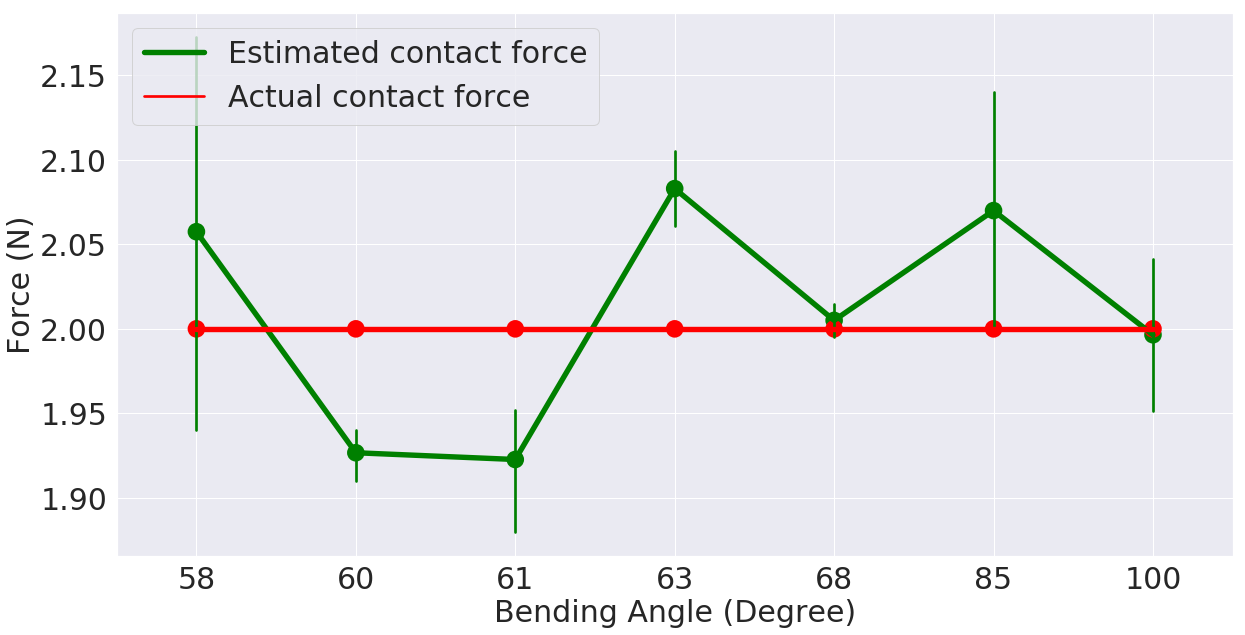}
  \end{center}
    \caption{Contact force estimation result of the force sensor.}
  \label{fig:estimation_result}
\end{figure}

The estimated contact force against the reference contact force at different bending angles is plotted in \figref{fig:estimation_result}. Although the graph shows that the estimate contact force varies slightly around the actual contact force, the error only ranges from 0.01 to 0.15 N at most, which is within the tolerance for the objects we will later grasp. All in all, estimating the contact force by simply subtracting the internal force from the real measurement is sufficiently accurate. 

\subsection{Force Control}
\label{sec:exp_force_controller}

Before conducting the experiment, we experimentally fine-tuned the proportional $K_p$ and integral $K_i$ gains of the force controller to 10 and 1.5, respectively. To experimentally evaluate the accuracy and stability of the force controller, each finger had to reach and track a variable reference signal. First, a finger was actuated with 65 kPa pressure to make contact with a solid object. Then, the step reference signal was increased from 0 to 3 N, and after 60 seconds the reference signal was set to 2 N. The sensory feedback from the embedded sensors was continuously fed to the internal force predictive model to estimate the actual contact force. The difference between the target and current contact force was then fed to the \ac{pi} controller in \eqref{eq:force_controller_discrete} whose output was the corresponding amount to be added to (or subtracted from) the current duty cycle signal. The experiment was repeated 5 times.
\begin{figure}[t]
  \begin{center}
    \includegraphics[width=0.48\textwidth]{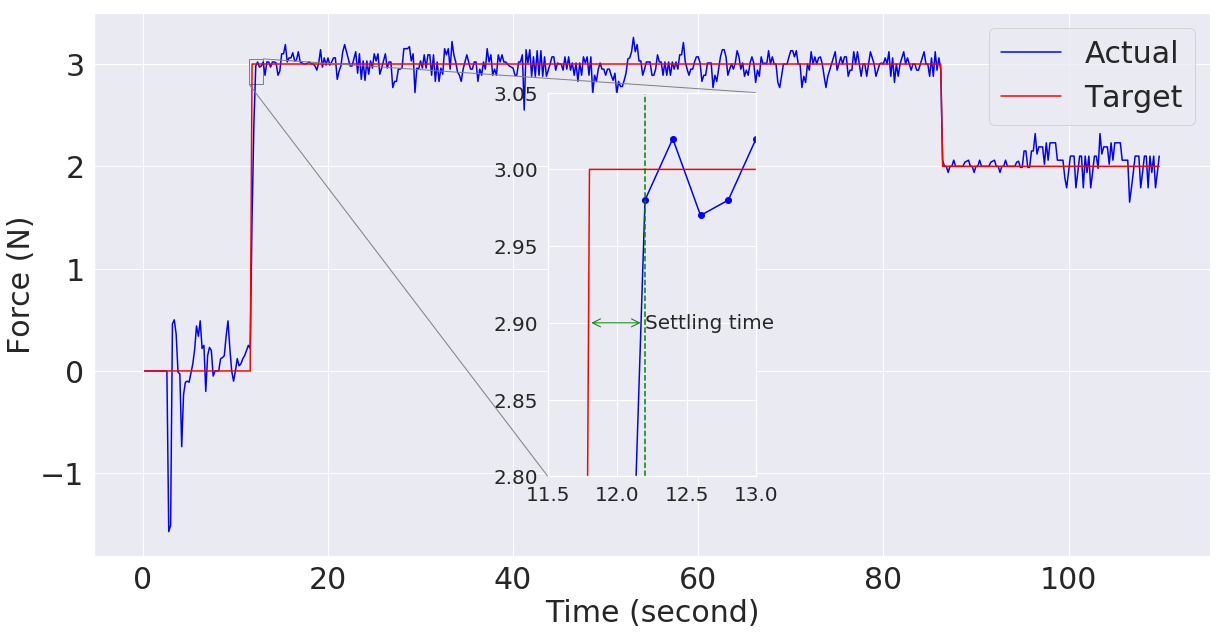}
  \end{center}
    \caption{Contact force response of the finger to step reference signals.}
  \label{fig:forcecontroller}
\end{figure}
The contact force response of one finger is shown in \figref{fig:forcecontroller}. The force response of the finger closely followed the step reference signals with a \ac{rms} error of 0.1 N. Moreover, the settling time of the finger was roughly 400 ms.  

To enhance the robustness of the grasping system, the force controller should only be activated when contact is detected. One option to achieve this is to utilize a switching mechanism between position and force control. We experimentally tested such a switching mechanism by having a finger reach a target contact force of 2.5 N only after contact was made. Initially, the finger was slowly actuated from a duty cycle of 0\% until contact with a solid object was detected based on the estimated contact force. After contact was established, the force controller was activated to regulate the pressure to achieve the target contact force.
\begin{figure}[t]
  \begin{center}
    \includegraphics[width=0.48\textwidth]{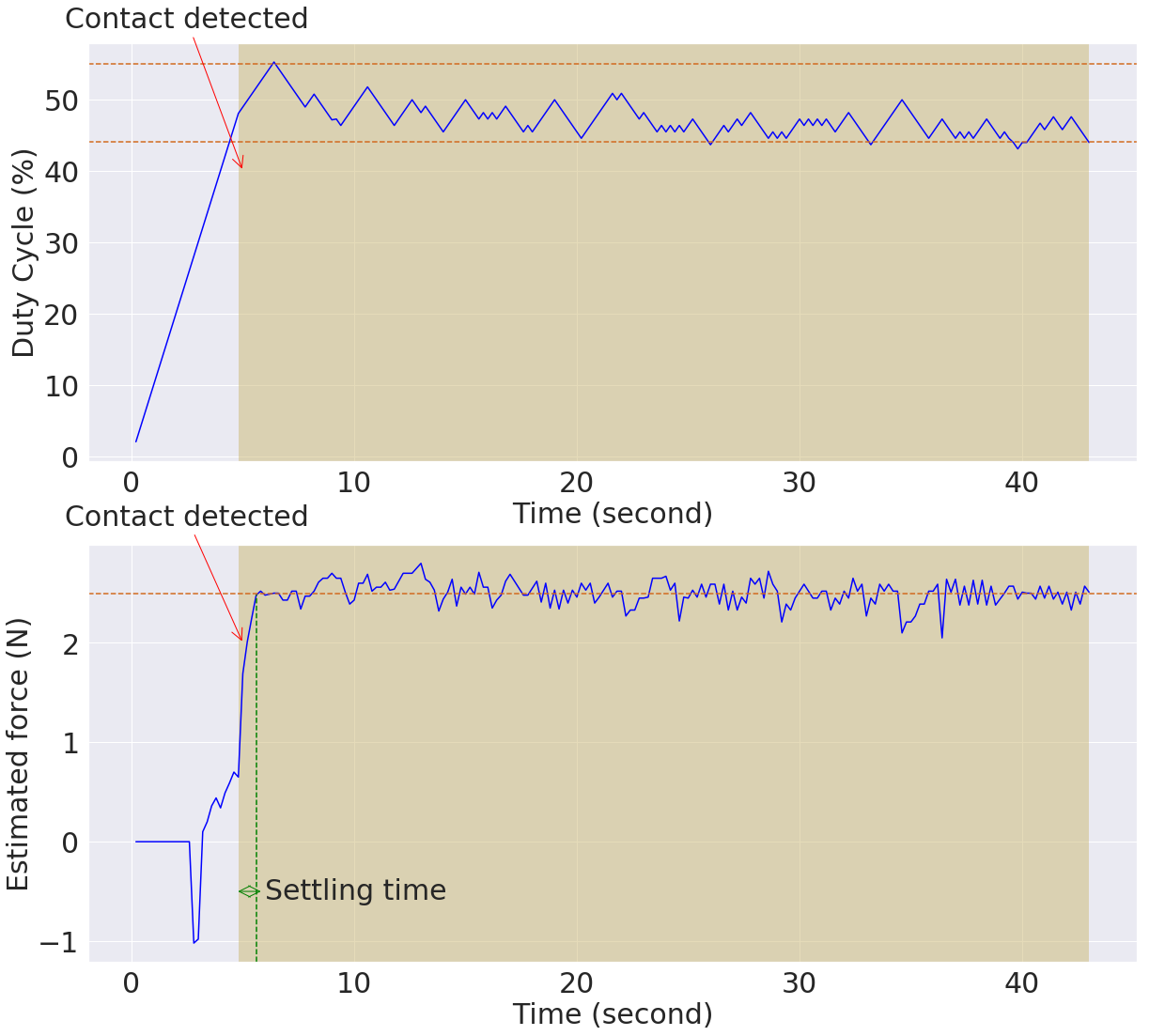}
  \end{center}
    \caption{The top figure shows the change in duty cycle to achieve the desired contact force while the bottom figure shows the contact force response of the force controller. The yellow region illustrates the duration the force controller was activated, the red dashed line represents the target contact force of 2.5 N and the green arrow indicates the settling time.}
  \label{fig:controlstrategy}
\end{figure}

\begin{table*}[htb]
    \centering
    \ra{1.1}
    \caption{The grasping results for the deformable objects\label{tb:final_results}.}
        \begin{tabular}{@{}ccccccccccccc@{}}
        \toprule
        \multirow{2}{*}{\makecell{Target contact force \\ of the opposable finger (N)}}
        & \multicolumn{2}{c}{Empty plastic cup} & \multicolumn{2}{c}{Empty paper cup} & \\
        \cmidrule(lr){2-3}\cmidrule(lr){4-5}
        & Dropped percentage{} & Deformed percentage{} & Dropped percentage{} & Deformed percentage{} &  \\
        \midrule
        4& 0\% & 100\% & 0\% & 100\%\\
        3& 0\% & 100\% & 0\% & 80\%\\
        {2}& 0\% & 90\% & {\textbf{0\%}} & {\textbf{20\%}}\\
        1.5& 0\% & 40\% & 30\% & 10\%\\
        {1}& {\textbf{0\%}} & {\textbf{10\%}} &  80\% &  0\%\\
        0.5& 60\% & 0\% & 100\% & 0\%\\
    \bottomrule
    \end{tabular}
\end{table*}

The duty cycle output from the controller and the contact force response are shown in \figref{fig:controlstrategy}. From the top figure, one can conclude that to achieve the target contact force the controller used a duty cycle value in the range of 45\% - 55\%. The main reason for the fluctuating duty cycle originates from fluctuations in the estimated contact force response caused by residual oscillations in the sensory reading. From the bottom figure, it is observed that the contact force response settled to a value of approximately 2.5 N in roughly 800 milliseconds. The \ac{rms} error between the measured and the target contact force was 0.21 N. In conclusion, regardless of the small fluctuations in the measured contact force and the output duty cycle, the proposed control strategy was successful in controlling the finger to achieve a target contact force in reasonable settling time. Furthermore, the proposed control strategy did not overshoot at the time of contact. 

Together, all these results confirm that it is possible to control the interaction force between the soft hand and objects in a safe, accurate, and fast manner. Such features are of uttermost importance in safe grasping. 

\subsection{Safe Grasping} 
\label{sec:safe_grasping}
\begin{figure}[t]
  \begin{center}
    \includegraphics[width=0.4\textwidth]{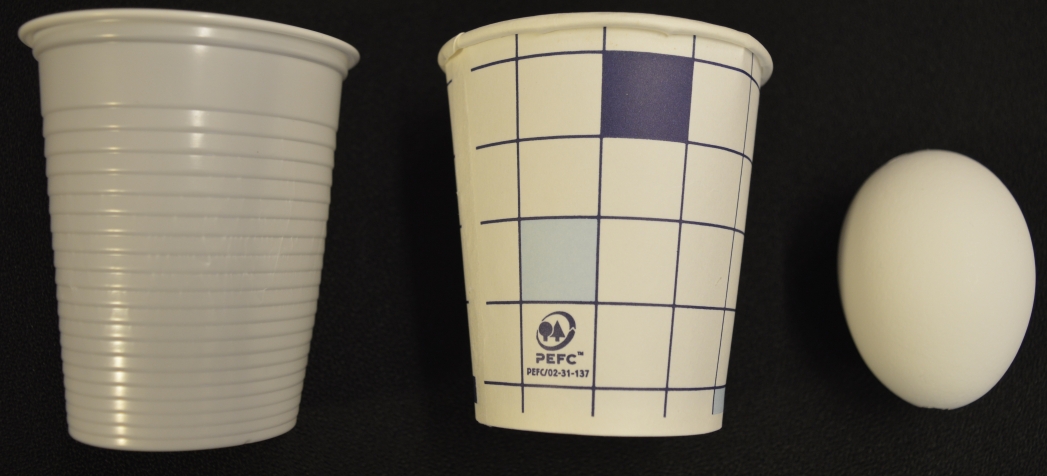}
  \end{center}
    \caption{The target objects used in the safe grasping experiment.  \label{fig:finalexp}
}
\end{figure}

Based on the results from the previous experiments, our proposed force controller exhibits characteristics for safe grasping. In safe grasping, the goal is to grasp deformable objects that can easily deform, such as the plastic cup seen in \figref{fig:resultfigure}, in a stable manner. We experimentally evaluated if the proposed force controlled soft hand can do safe grasping by grasping three sensitive objects: an empty plastic cup, an empty paper cup, and an empty eggshell all shown in \figref{fig:finalexp}. Of these objects, the empty plastic and paper cup represents deformable objects while the empty eggshell represents a fragile object.

For this experiment, the soft hand was fixed to a handle, as shown in \figref{fig:resultfigure}. As the soft hand had two fingers on one side and one on the opposite, to stabilize the objects the targets contact forces of the three fingers were set in such a way that the sum of the contact force of the two fingers was equal to that of the opposable finger. We tested six different force set-points for the opposable finger: $4$, $3$, $2$, $1.5$, $1$, and $0.5$ N. To evaluate a grasp, the hand first grasped the object until the target contact force was achieved and then we manually moved the handle upward 30 cm to lift the object and finally shook it randomly. During the process, a grasp was successful if the object neither deformed in the case of deformable objects nor broke in the case of fragile objects nor slipped away from the hand. We evaluated ten grasps for each target force amounting to 60 grasps in total. 

The grasping result on the deformable objects are presented in \tabref{tb:final_results}. As suspected, a higher contact force lowers the dropping rate but increases the deformation rate. The results show that even soft hands, which have been proposed for safe grasping due to their passive compliance, can completely deform objects if the force is too high. For instance, the plastic cup, which is less durable than the paper cup, can be deformed with as low a contact force as 1 N. The paper cup, on the other hand, can withstand up to 2 N of contact force. The minimum grasping force to successfully grasp deformable object can also be deduced from the results. 

In addition, the experiment was also conducted on an empty eggshell to evaluate the proposed control strategy in the case of fragile objects. In this case, the eggshell never broke not even with maximum target contact forces. However, with a soft hand that can generate much larger grasping forces, crushing fragile objects is indeed feasible. All in all, the results emphasize the need for integrating sensors into soft hands to manipulate deformable and fragile objects.

\subsection{Object properties estimation}
In addition to the previous experiments, we conducted one more extra experiment to show the potential of using the proposed soft hand to estimate object properties such as hardness. According to Yuan \cite{yuan16}, the most important factor to estimate the hardness of an object is the relationship between the geometry of the deformed object and the pressing force. When pressing on harder objects, they deform less compared to soft objects, thus retaining larger slopes on the contact surface \cite{yuan16}. In this work, the estimated contact force is seen as the pressing force, and the bending angle of the finger is considered as the deformation of the object. This experiment investigates whether the relationship between the two can be used to realize the hardness of an object.

\begin{figure}[t]
  \begin{center}
    \includegraphics[width=0.3\textwidth]{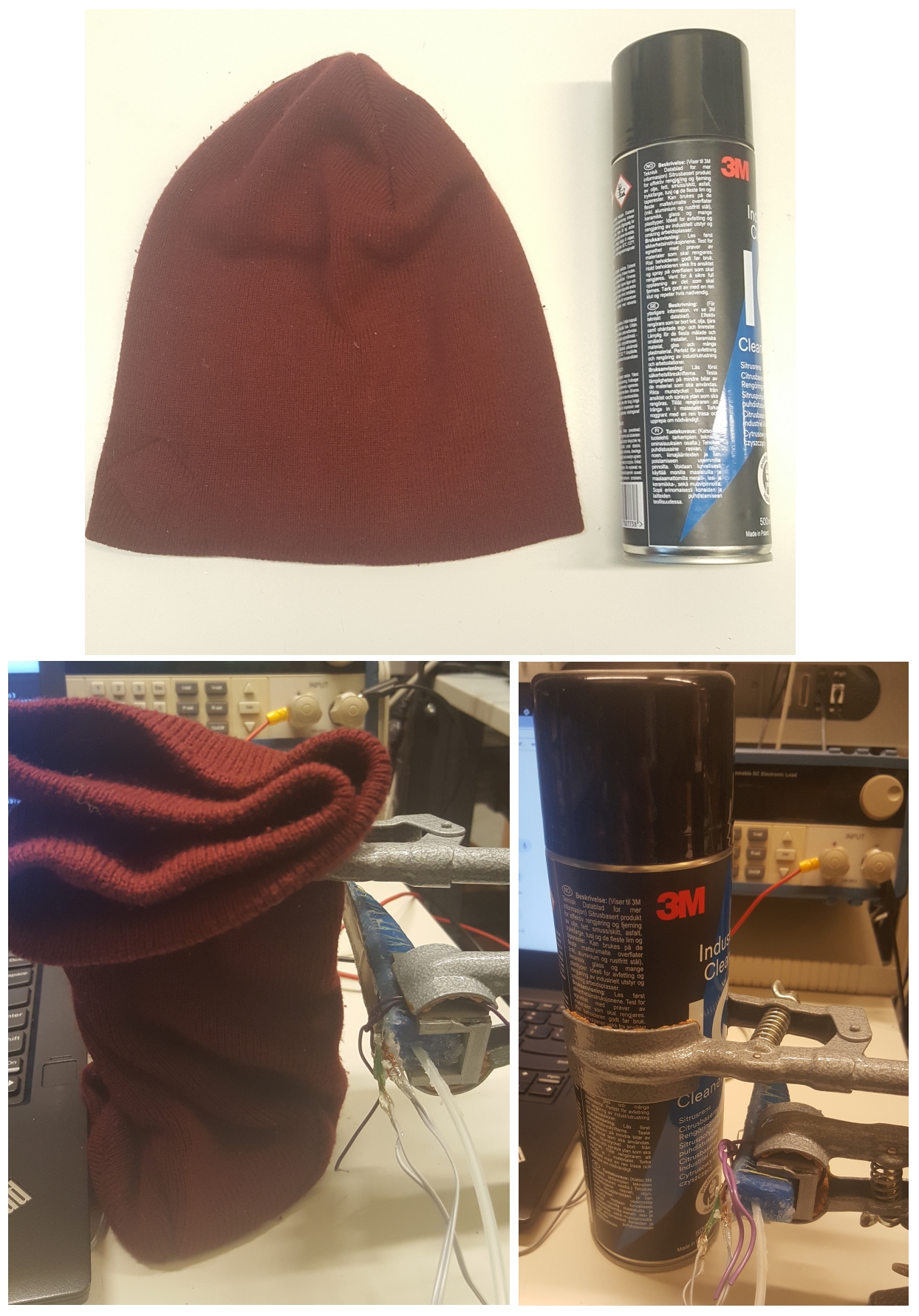}
  \end{center}
    \caption{The top figure shows the two target objects in this experiment: a woolly hat and a spray can. Bottom figures show the experimental setup for both objects.}
  \label{fig:softsolidobj}
\end{figure}

For this experiment, the finger was fixed to a handle and actuated to make contact with two objects with different hardness \ie a solid spray can, and a woolly hat as shown in \figref{fig:softsolidobj}. The target objects were placed in such a way that they will be in contact with the finger at 40\% duty cycle. Both force and position sensory readings were then recorded for analysis.

\begin{figure}[t]
  \begin{center}
    \includegraphics[width=0.5\textwidth]{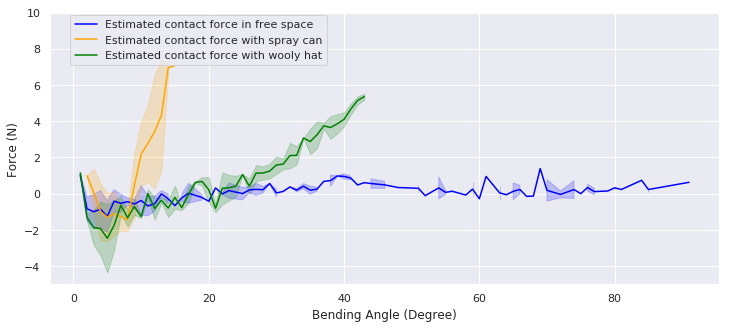}
  \end{center}
    \caption{The blue line represents the estimated contact force when the finger bends in free space. The orange and green line represent the estimated contact force when the finger makes contact with the spray can and the woolly hat, respectively.}
  \label{fig:softvssolidbend}
\end{figure}

\figref{fig:softvssolidbend} plots the relationship between the bending angle and the estimated contact force in both cases. The results show that in the case of the spray can the bend angle remains almost constant while the contact force continues to increase. This means that the finger has been stopped by something stiff. And since the finger is kept actuating, it keeps pressing stronger against that stiff object resulting in the increase of the contact force. However, in the case of the woolly hat, both the bending angle and the contact force increase simultaneously after the contact. This indicates that the target object is not stiff enough to constrain the bending of the finger after contact. Based on these results, it seems that the soft finger embedded with selected sensors can successfully distinguish between a solid object and a soft object using only the sensory feedback. 
\section{CONCLUSIONS}
\label{ch:Conclusion}

We presented a force controlled soft robotic hand and used it for safe grasping. The key component was to integrate both resistive bend and force sensors onto the hand's fingers and apply a data-driven method to estimate the actual contact force between the fingers and the objects. The estimated contact force was then fed into a \ac{pi} force controller to keep a constant grasping force. We experimentally validated our force controller by comparing it to no force controller on a grasping experiment where the objective was to grasp deformable objects in a stable manner without damaging them. The results show that the force controlled soft hand could grasp the tested objects without neither dropping them nor causing significant deformation.

All in all, the work presented here demonstrates that applying a data-driven calibration method can make otherwise unreliable force sensor readings reliable enough to be used in a force feedback controller. This, in turn, poses new interesting research questions. For instance, is it possible to learn an accurate dynamics model from the force and bend data? If so, can such a model be used to do dexterous manipulation by a soft robotic hand with, \eg{} model predictive control or model-based reinforcement learning? Dexterous manipulation, together with the safety inherited from the softness properties of soft hands offer great potential for robots to interact with human in complex environments. 
\bibliographystyle{IEEEtran}
\bibliography{refs}

\end{document}